\documentclass{article}

    \usepackage[preprint]{neurips_2020}

\usepackage{microtype}
\usepackage{graphicx}
\usepackage{subfigure}
\usepackage{booktabs} 

\usepackage{amsmath,amsfonts,bm}

\def\eqref#1{equation~\ref{#1}}

\def\1{\bm{1}}

\DeclareMathAlphabet{\mathsfit}{\encodingdefault}{\sfdefault}{m}{sl}
\SetMathAlphabet{\mathsfit}{bold}{\encodingdefault}{\sfdefault}{bx}{n}

\def\gT{{\mathcal{T}}}

\usepackage{balance}
\usepackage{hyperref}
\usepackage{url}
\usepackage[utf8]{inputenc} 
\usepackage[T1]{fontenc}    
\usepackage{hyperref}       
\usepackage{url}            
\usepackage{booktabs}       
\usepackage{amsfonts}       
\usepackage{nicefrac}       
\usepackage{microtype}      
\usepackage{graphicx}
\usepackage{amsmath} 
\usepackage{amssymb}
\usepackage{enumitem,kantlipsum}
\usepackage{float}
\usepackage[export]{adjustbox}
\usepackage{subfigure}
\usepackage{caption}
\usepackage{multirow}
\usepackage[ruled]{algorithm2e}
\usepackage{algpseudocode}
\usepackage{array}
\newcolumntype{P}[1]{>{\centering\arraybackslash}p{#1}}
\usepackage{textcomp}
\usepackage{booktabs}
\usepackage{makecell}
\usepackage{todonotes}
\usepackage{kantlipsum}
\usepackage{mdwlist}
\usepackage{wrapfig}

\setcitestyle{round}
\newcommand{\bb}[1]{\mathbf{#1}}

\newcommand{\bx}{\bb{x}}

\newcommand{\by}{\bb{y}}

\newcommand{\bz}{\bb{z}}

\newcommand{\bbf}{\bb{f}}

\newcommand{\bT}{\boldsymbol{\theta}}

\newcommand{\pT}{p_{\bT}}

\newcommand{\fT}{\bbf_{\bT}}

\DeclareMathOperator{\tr}{tr}

\usepackage{hyperref}

\title{Likelihood Contribution based Multi-scale Architecture for Generative Flows}

\author{%
Hari Prasanna Das, Pieter Abbeel and Costas J. Spanos
  \\
  Department of Electrical Engineering and Computer Sciences\\
  University of California, Berkeley\\
  \texttt{\{hpdas,pabbeel,spanos\}@berkeley.edu} \\
}

\begin{document}

\maketitle
\begin{abstract}
Deep generative modeling using flows has gained popularity owing to the tractable exact log-likelihood estimation with efficient training and synthesis process. However, flow models suffer from the challenge of having high dimensional latent space, the same in dimension as the input space. An effective solution to the above challenge as proposed by \citet{dinh2016real} is a multi-scale architecture, which is based on iterative early factorization of a part of the total dimensions at regular intervals. Prior works on generative flow models involving a multi-scale architecture perform the dimension factorization based on static masking. We propose a novel multi-scale architecture that performs data-dependent factorization to decide which dimensions should pass through more flow layers. To facilitate the same, we introduce a heuristic based on the contribution of each dimension to the total log-likelihood which encodes the importance of the dimensions. Our proposed heuristic is readily obtained as part of the flow training process, enabling the versatile implementation of our likelihood contribution based multi-scale architecture for generic flow models. We present such implementations for several state-of-the-art flow models and demonstrate improvements in log-likelihood score and sampling quality on standard image benchmarks. We also conduct ablation studies to compare the proposed method with other options for dimension factorization.
\end{abstract}
\section{Introduction}
Deep Generative Modeling aims to learn the embedded distributions and representations in input (especially unlabelled) data, requiring no/minimal human labelling effort. 
The representations learnt can then be utilized in a number of downstream tasks such as semi-supervised learning \citep{kingma2014semi,odena2016semi}, synthetic data augmentation and adversarial training \citep{cisse2017parseval}, text analysis and model based control etc. The repository of deep generative modeling majorly includes likelihood based models such as autoregressive models, latent variable models, flow based models and implicit models such as generative adversarial networks (GANs). Autoregressive models \citep{salimans2017pixelcnn++,oord2016pixel,van2016conditional,chen2017pixelsnail} achieve exceptional log-likelihood score on many standard datasets, indicative of their power to model the inherent distribution. But, they suffer from a slow sampling process. Latent variable models such as variational autoencoders \citep{kingma2013auto} tend to better capture the global feature representation in data, but do not offer an exact density estimate. Implicit generative models such as GANs optimize a generator and a discriminator in a min-max fashion have recently become popular for their ability to synthesize realistic data \citep{karras2018style,engel2019gansynth}. But, GANs neither offer a latent space suitable for further downstream tasks, nor do they perform density estimation. 

Flow based generative models \citep{dinh2016real,kingma2018glow} perform exact density estimation with fast inference and sampling, due to their parallelizability. They also provide an information rich latent space suitable for many applications. However, the dimension of latent space for flow based generative models is same as the high-dimensional input space, by virtue of bijective nature of flows. This poses a bottleneck for flow models to scale with increasing input dimensions due to computational complexity. An effective solution to the above problem is a multi-scale architecture, introduced by \citet{dinh2016real}, which performs iterative early gaussianization of a part of the total dimensions at regular intervals of flow layers. This not only makes the model computational and memory efficient but also aids in distributing the loss function throughout the network for better training. Many prior works including \citet{kingma2018glow,behrmann2018invertible,atanov2019semi,kumar2019videoflow} implement multi-scale architecture in their flow models, but use static masking methods for factorization of dimensions. 

We propose a multi-scale architecture which performs data dependent factorization to decide which dimensions should pass through more flow layers. For the decision making, we introduce a heuristic based on the amount of log-likelihood contribution by each dimension, which in turn signifies their individual importance. Since in the proposed architecture, the heuristic is obtained as part of the flow training process, it can be universally applied to generic flow models. We present such implementations for flow models based on affine/additive coupling as well as ordinary differential equations (ODE) and achieve quantitative and qualitative improvements. We also perform ablation studies to establish the novelty of our method. Summing up, the contributions of our research are,

\textbf{Contributions:} 
\vspace{-2mm}
\begin{enumerate}
\item 
A log-determinant based heuristic which entails the contribution by each dimension towards the total log-likelihood in a multi-scale architecture.
\item
A multi-scale architecture based on the above heuristic performing data-dependent splitting of dimensions, implemented for several classes of flow models.
\item
Quantitative and qualitative analysis of above implementations and an ablation study
\end{enumerate}
 To the best of our knowledge, we are the first to propose a data-dependent splitting of dimensions in a multi-scale architecture.

\section{Background}
\vspace{-2mm}
\subsection{Flow-based Generative Models}\label{background:flow}
\vspace{-2mm}
Let $\bx$ be a high-dimensional random vector with unknown true distribution $p(\bx)$. The following formulation is directly applicable to continous data, and with some pre-processing steps such as dequantization \citep{uria2013rnade,salimans2017pixelcnn++,ho2019flow++} to discrete data. Let $\bz$ be the latent variable with a known standard distribution $p(\bz)$, such as a standard multivariate gaussian. Using an i.i.d. dataset $\mathcal{D}$, the target is to model $\pT(\bx)$ with parameters $\bT$. A flow, $\fT$ is defined to be an invertible transformation that maps observed data $\bx$ to the latent variable $\bz$. A flow is invertible, so the inverse function $\gT$ maps $\bz$ to $\bx$, i.e.
\begin{align}
    \bz = \fT(\bx) = \gT^{-1}(\bx)\;\;\text{and}\;\;
    \bx = \gT(\bz) = \fT^{-1}(\bz)
\end{align}
The log-likelihood can be expressed as,
\begin{align}
\log\pT(\bx) &= \log p(\bz) + \log\left|\det\left(\frac{\partial \fT(\bx)}{\partial \bx^{T}}\right)\right|\label{eq:flow_func_repr}
\end{align}
where $\cfrac{\partial \fT(\bx)}{\partial \bx^{T}}$ is the Jacobian of $\fT$ at $\bx$. The invertible nature of a flow allows it to be capable of being composed of other flows of compatible dimensions. In practice, flows are constructed by composing a series of component flows. Let the flow $\fT$ be composed of $K$ component flows, i.e. $\fT = \bbf_{\theta_{K}} \circ \bbf_{\theta_{K-1}} \circ \cdots \circ \bbf_{\theta_1}$ and the intermediate variables be denoted by $\by_K,\by_{K-1},\cdots,\by_{0}=\bx$. Then the log-likelihood of the composed flow is,
\begin{align}
\log\pT(\bx) &= \underbrace{\log p(\bz)}_{\text{Log-latent density}} + \underbrace{\log\left|\det\left(\frac{\partial (\bbf_{\theta_K} \circ \bbf_{\theta_{K-1}} \circ \cdots \circ \bbf_{\theta_1}(\bx))}{\partial \bx^{T}}\right)\right|}_{=\sum_{i=1}^{K}\log|\det(\partial\by_i/\partial\by_{i-1}^T)| \;\;\text{(Log-det)}}\label{eq:log_lkd}
\end{align}
which follows from the fact that $\det(A\cdot B) = \det(A)\cdot \det(B)$. In our work, we refer the first term in Equation \ref{eq:log_lkd} as \textit{log-latent-density} and the second term as \textit{log-determinant (log-det)}. The reverse path, from $\bz$ to $\bx$ can be written as a composition of inverse flows, 
$\bx = \fT^{-1}(\bz) = \bbf_{\theta_{1}}^{-1} \circ \bbf_{\theta_{2}}^{-1} \circ \cdots \circ \bbf_{\theta_K}^{-1}(\bz)$. Confirming with above properties for a flow, different types of flows can be constructed \citep{kingma2018glow,dinh2016real,dinh2014nice, behrmann2018invertible,chen2019residual}. 
\vspace{-3mm}
\subsection{Multi-scale Architecture}\label{background:ma}
\vspace{-2mm}
Multi-scale architecture is a design choice for latent space dimensionality reduction of flow models, in which part of the dimensions are factored out/early gaussianized at regular intervals, and the other part is exposed to more flow layers. The process is called dimension factorization. In the problem setting as introduced in Section~\ref{background:flow}, the factoring operation can be mathematically expressed as,
\begin{align*}
    \by_{0} = \bx, 
    (\bz_{l+1},\by_{l+1}) &= \bbf_{\theta_{l+1}}(\by_{l}),\;\;l \in \{0,1,\cdots,K-2\}\label{eq:recur_fact}\\
    \bz_K = \bbf_{\theta_{K}}(\by_{K-1}),
    \bz &= (\bz_{1},\bz_{2},\cdots,\bz_{K})
\end{align*}
The factoring of dimensions at early layers has the benefit of distributing the loss function throughout the network \citep{dinh2016real} and optimizing the amount of computation used by the model. 

\section{Likelihood Contribution based Multi-scale Architecture}\label{sensitive}
In a multi-scale architecture, it is apparent that the network will better learn the distribution of dimensions getting exposed to more layers of flow as compared to the ones which get factored at a finer scale (earlier layer). The method of dimension splitting proposed by prior works such as \citep{dinh2016real,kingma2018glow,behrmann2018invertible} are static in nature and do not distinguish between importance of different dimensions. In this section, we introduce the general framework for likelihood contribution based heuristic and associated multi-scale architecture along with its integration with flow training process.
\vspace{-3mm}
\subsection{Likelihood Contribution based Heuristic}
Recall from Equation \ref{eq:log_lkd} that the log-likelihood is composed of two terms, the log-latent density term and the log-det term. We focus on the log-det term since it depends on the modeling of flow layers. 

Let the dimension of the input (images in our case) space $\bx$ be $s\times s\times c$, where $s$ is the height/width and $c$ is the number of channels. For the following formulation, let us assume NO dimensions were gaussianized early so that we have access to log-det term for all dimensions at each flow layer, and the dimension at all intermediate layers remain the same (i.e. $s\times s\times c$). We apply a flow ($\fT$) with $K$ component flows to $\bx$, $\bz$ pair, so that $\bz = \fT(\bx) = \bbf_{\theta_{K}} \circ \bbf_{\theta_{K-1}} \circ \cdots \circ \bbf_{\theta_1}(\bx)$. The intermediate variables are denoted by $\by_K,\by_{K-1},\cdots,\by_{0}$ with $\by_{K} = \bz$ (since no early gaussianization was performed) and $\by_{0} = \bx$.
The log-det term at layer $l$, $L_d^{(l)}$, is given by,
\begin{align}
    [L_d^{(l)}]_{scaler} &= \sum_{i=1}^{l} \log|\det(\partial\by_i/\partial\by_{i-1}^T)|
\end{align}
The log-det of the jacobian term encompasses contribution by all the $s\times s\times c$ dimensions. If we decompose it to obtain the individual contribution by the dimensions (we discuss explicitly on how to perform this decomposition in Sec.~\ref{sec:dim-decomposition}) towards the total log-det ($\sim$ total log-likelihood). The log-det term can be viewed (with slight abuse of notations) as a $s\times s\times c$ tensor corresponding to each of the dimensions, summed over the flow layers till $l$,
\begin{align*}
\vspace{-8mm}
    [L_d^{(l)}]_{s\times s\times c} &= \sum_{i=1}^{l} [d_{i-1}^{(\alpha,\beta,\gamma)}]_{s\times s\times c},\;\;\\\text{where}\;\;\alpha,\beta \in\{0,\cdots,s\}\;\; \text{and}\;\; \gamma\in\{0,\cdots,&c\},
    \;\;\text{s.t.}\; \sum_{\alpha,\beta,\gamma}d_{i-1}^{(\alpha,\beta,\gamma)} = \log|\det(\partial\by_i/\partial\by_{i-1}^T)|
    \vspace{-8mm}
\end{align*}
The entries in $[L_d^{(l)}]_{s\times s\times c}$ having higher value were scaled up more, and correspond to dimensions which are more sensitive to changes in input, so the flow can learn more by processing them through more layers. So, we can use the \textit{likelihood contribution (in the form of log-det term) by each dimension} as a heuristic for deciding which variables should be gaussianized early in a multi-scale architecture.

\subsection{Estimation of Per-Dimensional Likelihood Contribution for various types of Flows}\label{sec:dim-decomposition}
The likelihood (log-det) contributions per dimension can be obtained by decomposition of the overall log-det of the jacobian. Now, we describe the decomposition process for various types of flow models. The log-det per dimension after decomposition is averaged across the samples in training set, so as to obtain an overall representative likelihood contribution by each dimension.
\vspace{-4mm}
\subsubsection{Affine coupling based flows}\label{sec:affine_additive}
\textit{RealNVP \citep{dinh2016real}:} For RealNVP with affine coupling layers, the logarithm of individual diagonal elements of jacobian, summed over layers till $l$ provides the per-dimensional likelihood contribution components at layer $l$.

\textit{Glow \citep{kingma2018glow}:} Unlike RealNVP where the log-det terms for each dimension can be expressed as log of corresponding diagonal element of jacobian, Glow contains $1\times 1$ convolution blocks having non-diagonal log-det term. For a $s\times s\times c$ tensor, the log-det term is $s \cdot s \cdot \log | \det(W) |$, where $W$ is the weight matrix.
It is clear that the contribution by a pixel is $\log | \det(W) |$, and it has to be decomposed to obtain individual contribution by each channel. As a suitable candidate, singular values of $W$ correspond to the contribution from each channel dimension, hence their log value is the individual log-det contribution. The individual log-det term for channels are obtained by,
\begin{align}
  | \det(W) | &= \prod_{i}\sigma_{i}(W) \Leftrightarrow \log | \det(W) | = \sum_{i}\log(\sigma_{i}(W))
\end{align}
where $\sigma_{i}(W)$ are the singular values of $W$. For affine blocks, same method as RealNVP is adopted.
\subsubsection{Flow models with ODE based Density Estimators} \label{sec:ode}
Recent works on flow models such as \citet{behrmann2018invertible,grathwohl2018ffjord,chen2019residual} employ variants of ODE based density estimators. The following formulation is applicable to find per-dimensional likelihood contributions for such flow models. In the above works, the flow is modelled as $F(x)$, such that $z= F(x) = (I + g) (x)$, where $g(x)$ is the forward propagation function. The log-det of the jacobian is expressed as a power series,
\begin{align*}
   \ln |\det J_F(x)| = \tr\left( \ln \big(I+ J_g (x)\big)\right)= \sum_{k=1}^\infty (-1)^{k+1}\frac{\tr(J_g^k)}{k}
\end{align*}
where $\tr$ denotes the trace. Due to computational constraints, the power series is computed up to a finite number of iterations with the $\tr(J_g^k)$ term stochastically approximated by hutchinson's trace estimator, $\tr(A) = \mathbb{E}_{p(v)} \left[v^T A v \right]$, with  $\mathbb{E}[v] = 0$ and $\text{Cov}(v) = I$. The component corresponding to each dimension that becomes part of the log-det term is the diagonal element of $J_g^k$, so the per-dimension contribution to the likelihood can be approximated as the diagonal elements of $J_g^k$, summed over the power series upto a finite number of iterations $n$. The diagonal elements are obtained with the hutchinson's trace estimator without any extra cost, i.e. if $v = [v_1,v_2,\cdots,v_{s\times s\times c}]^T$,
\begin{align*}
   \sum_{k=1}^\infty (-1)^{k+1}\frac{\tr(J_g^k)}{k} &= \sum_{k=1}^\infty (-1)^{k+1}\frac{\mathbb{E}_{p(v)} \left[v^T J_g^k v \right]}{k} = 
   \sum_{k=1}^\infty (-1)^{k+1}\frac{\mathbb{E}_{p(v)} \left[(v^T J_g^k) v \right]}{k}
\end{align*}
In the above equation, $(v^T J_g^k)$ is the vector-jacobian product which is multiplied again with $v$. The individual components which are summed when $(v^T J_g^k)$ is multiplied with $v$ correspond to the diagonal terms in jacobian, over the expectation $\mathbb{E}_{p(v)}$. So those terms are the contribution by the individual dimensions to the log-likelihood and are expressed as $[L_d^{(l)}]_{s\times s\times c}$ at flow layer $l$.
\begin{algorithm}[tb]
  \caption{LCMA Implementation and Training for Generative Flow models}
  \label{alg:lcma_realnvp}
  {\bfseries Pre-training Phase:} Pre-train a flow model with no multi-scale architecture (no dimensionality reduction) to obtain the log-det terms ($[L_d^{(l)}]_{s\times s\times c}$) at each layer $l$\\
  
  {\bfseries Dimension Factorization Phase:} Initialize the input dimensions $\phi\times\phi\times\psi$ $\xleftarrow{} s\times s\times c$\\
  \While{$1\leq l\leq K$}{
    Select $[L_d^{(l)}]_{\phi\times\phi\times\psi}$ corresponding to input dimensions.\\
    $[L_d^{(l)}]_{\phi\times\phi\times\psi}$ $\xrightarrow[\text{(Figure 1)}]{\text{Local Max- \& Min-Pooling}}$ $[L_d^{(l)}]_{\frac{\phi}{2}\times \frac{\phi}{2}\times 4\psi}$ $\xrightarrow[\text{Splitting}]{\text{Channelwise}}$ $[L_d^{(l, \text{Max})},L_d^{(l, \text{Min})}]_{\frac{\phi}{2}\times \frac{\phi}{2}\times 2\psi}$\\
    Gaussianize the dimensions corresponding to $L_d^{(l, \text{Min})}$ and pass the dimensions corresponding to $L_d^{(l, \text{Max})}$ to more flow layers\\
    $\phi\times\phi\times\psi$ $\xleftarrow{}\frac{\phi}{2}\times \frac{\phi}{2}\times 2\psi$
  }
  {\bfseries Training Phase:} Flow model with proposed LCMA is trained using maximum likelihood. 
\end{algorithm}

\subsection{Dimension Factorization using Proposed Heuristic}\label{sec:dim-fact}
At every flow layer, an ideal factorization method should,
\begin{enumerate}[labelindent=0pt]
    \item{\textit{(Quantitative) For efficient density estimation:} Early gaussianize the dimensions having less log-det and expose the ones having more log-det to more flow layers. In this manner, selectively more expressivity can be provided to dimensions which capture meaningful representations (and are more valuable from a log-det perspective).}
    \item{\textit{(Qualitative) For qualitative reconstruction:} Capture the local variance over the flow layers, i.e. the dimensions being exposed to more flow layers should contain representative pixel variables from throughout the whole image.}
\end{enumerate}


We perform a hybrid dimension factorization taking both of the above criterias into account. A $s\times s\times c$ tensor is converted to $\frac{s}{2}\times\frac{s}{2}\times 4c$ tensor, by using local max and min pooling operations on corresponding per dimensional log-det terms as obtained in Sec.~\ref{sec:dim-decomposition} (which was averaged over the training set, so the learned splitting remains same for all data points). Then the $\frac{s}{2}\times\frac{s}{2}\times 4c$ tensor is split along channel \begin{wrapfigure}[16]{r}{0.38\textwidth}
\vspace{-4mm}
  \begin{center}
    \includegraphics[width=0.36\textwidth]{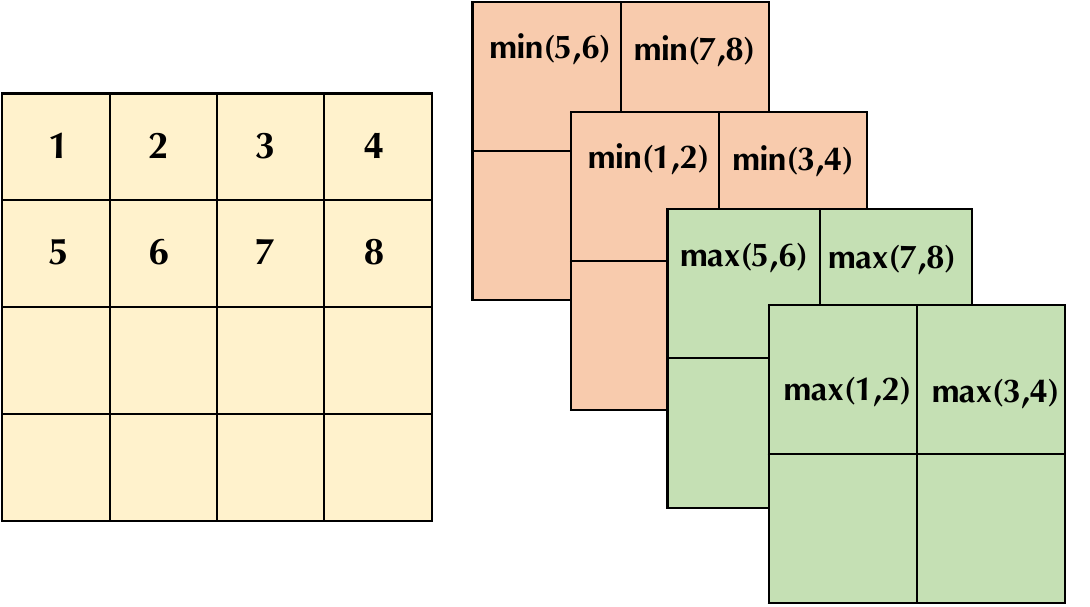}
  \end{center}
  \caption{Likelihood contribution based squeezing operation: (On left) The tensor $[L_d^{(l)}]_{s\times s\times c}$ representing log-det of variables in a flow layer. (On right) It is squeezed to $\frac{s}{2}\times \frac{s}{2}\times 4c$ with local max and min pooling operation. The green (orange) marked pixels represent dimensions having more (less) log-det locally.}
  \label{fig:sensitive_squeezing}
\end{wrapfigure}dimension to form two $\frac{s}{2}\times\frac{s}{2}\times 2c$ tensors, one corresponding to low log-det dimensions and one corresponding to high ones. The local min and max pooling operations (illustrated in Fig ~\ref{fig:sensitive_squeezing}) preserve the local spatial variation of the image in both parts of the factorization, leveraging both enhanced density estimation as well as qualitative reconstruction. 

\textbf{Factorization for Flows involving Squeezing Operation:} If squeezing operation (which is nothing but reordering of dimensions) is involved, we keep track of which dimensions belong to the half that get gaussianized early and which dimensions belong to the other half that passes through more flow layers. At the next layer, only the log-det terms for the dimensions which came through flow layers are considered for further splitting operation. 

We refer to the multi-scale architecture obtained by the above data-dependent dimension splitting method as Likelihood Contribution based Multi-scale Architecture (LCMA).

\subsection{Integration of LCMA with Flow Model Training}
Training of flow models with LCMA implementation is summarized in Algorithm \ref{alg:lcma_realnvp}.\\
\textbf{Pre-training Phase} involves training a non-multi-scale (NMS) flow model to obtain the individual contribution of dimensions ($[L_d^{(l)}]_{s\times s\times c}$) towards the total log-likelihood (Section ~\ref{sec:dim-decomposition}). Training a NMS model is computation heavy, but better training leads to improved density estimation score for resulting LCMA. So, there remains a trade-off between the amount of NMS pre-training and the density estimation performance for resulting LCMA. We train the NMS model partially (we report the results for varying levels of NMS pre-training and resulting density estimation performance for one of the flow training in the supplementary materials).

\textbf{Dimension Factorization Phase} involves deciding the dimensions to be factored out at each flow layer using proposed log-det heuristic. The dimension splittings are decided recursively for all the flow layers based on their likelihood contribution obtained in the previous phase.

\textbf{Training Phase} is the final stage where the flow model with LCMA is trained. Since the decision for factorization of dimensions at each flow layer occurs before the training starts, and the decision remains unchanged during training, change of variables formula can be applied. In fact, this allows the use of non-invertible operations (e.g. max and min pooling) for efficient factorization (Sec.~\ref{sec:dim-fact}).

\textbf{Time Complexity:} The time complexity associated with original multi-scale architecture is not affected by LCMA implementation as no additional time consuming blocks were added. Our data-dependent dimension splitting operation can be interpreted as replacing the conventional checkerboard/channel split masking with likelihood contribution based masking.

\section{Related Work}
\vspace{-3mm}
For multi-scale architectures in generative flow models, our proposed method performs factorization of dimensions based on their likelihood contribution, which in another sense translates to determining which dimensions are important from density estimation and qualitative reconstruction point of view. Keeping this in mind, we discuss prior works on generative flow models which involve multi-scaling and/or incorporate permutation among dimensions to capture their interactions. 

A number of generative flow models implement a multi-scale architecture, such as \citet{dinh2016real,kingma2018glow,atanov2019semi,izmailovsemi,durkan2019cubic,behrmann2018invertible,chen2019residual} etc.  \citet{kingma2018glow} introduce an $1\times 1$ convolution layer in between the actnorm and affine coupling layer in their flow architecture. The $1\times 1$ convolution is a generalization of permutation operation which ensures that each dimension can affect every other dimension. This can be interpreted as redistributing the contribution of dimensions to total likelihood among the whole space of dimensions. So \citet{kingma2018glow} treat the dimensions as equiprobable for factorization in their implementation of multi-scale architecture, and split the tensor at each flow layer evenly along the channel dimension. We, on the other hand, take the next step and focus on the individuality of dimensions and their importance from the amount they contribute towards the total log-likelihood. The log-det score is available via direct/indirect decomposition of the jacobian obtained as part of computations in a flow training, so we essentially have an easily available heuristic. Since LCMA focuses individually on the dimensions using easily available heuristic, it can prove to be versatile in being compatible with generic multi-scale architectures. \citet{hoogeboom2019emerging} extend the concept of $1\times 1$ convolutions to invertible $d\times d$ convolutions, but do not discuss multi-scaling. \citet{dinh2016real} also mention a type of permutation which is equivalent to reversing the ordering of the channels, but is more restrictive and fixed. 

Flow models such as \citet{behrmann2018invertible,grathwohl2018ffjord,chen2019residual} involve ODE based density estimators. They also implement a multi-scale architecture, but the splitting operation is a static channel wise splitting without considering the importance of individual dimensions or any permutations. ~\citet{izmailovsemi, durkan2019cubic, kumar2019videoflow,atanov2019semi} use multi-scale architecture in their flow models, coherent with ~\citet{dinh2016real,kingma2018glow}, but still perform the factorization of dimensions using static masking. For qualitative sampling along with efficient density estimation, we also propose that factorization methods should preserve spatiality of the image in the two splits, motivated by the spatial nature of splitting methods in \citet{kingma2018glow} (channel-wise splitting) and \citet{dinh2016real} (checkerboard and channel-wise splitting). Summarizing, we propose a data-dependent approach to dimension factorization in a multi-scale architecture, unexplored by prior works. 
\vspace{-4mm}

\section{Experiments}\label{results:likelihood}
In this section we present the detailed results of proposed LCMA adopted for the flow model of RealNVP \citep{dinh2016real}, Glow \citep{kingma2018glow},i-ResNet \citep{behrmann2018invertible} and Residual Flows \citep{chen2019residual}. For direct comparison, all the experimental settings such as data pre-processing, optimizer parameters as well as flow architectural details (coupling layers, residual blocks) are kept the same, only the factorization of dimensions at each flow layer is performed as per LCMA. The computations were performed in NVIDIA Tesla V100 GPUs. For RealNVP, we perform experiments on four benchmarked image datasets: \emph{CIFAR-10} \citep{krizhevsky2009learning}, \emph{Imagenet} \citep{russakovsky2014imagenet} (downsampled to $32\times 32$ and $64\times 64$), and \emph{CelebFaces Attributes (CelebA)} \citep{liu2015deep}. The scaling in LCMA is performed once for CIFAR-10, thrice for Imagenet $32\times 32$ and 4 times for Imagenet $64\times 64$ and CelebA. We compare LCMA with conventional RealNVP and report the quantitative and qualitative results. For Glow, i-ResNet and Residual Flows with LCMA, we perform experiments on CIFAR-10 and report improvements over baseline bits/dim (BPD).

\subsection{Quantitative Comparison}
\begin{wraptable}[6]{r}{0.53\textwidth}
\vspace{-4mm}
    \caption{Improvements in density estimation (in bits/dim) using proposed method for RealNVP}
    \vspace{-4mm}
    \label{table:log_likelihood}
    \begin{center}
    \resizebox{0.52\textwidth}{!}{%
    \begin{tabular}{p{3cm} | p{1cm} | p{1.5cm} | p{2.3cm} | p{2.3cm}}
    \toprule
    \centering
    Model     & CelebA & CIFAR-10 & ImageNet 32x32 & ImageNet 64x64\\
    \midrule
    \centering
    RealNVP & \makecell{$3.02$} & \makecell{$3.49$} & \makecell{$4.28$} & \makecell{$3.98$}\\
    \midrule
    \centering
    RealNVP with LCMA   & \makecell{$\mathbf{2.71}$} & \makecell{$\mathbf{3.43}$} & \makecell{$\mathbf{4.21}$} & \makecell{$\mathbf{3.92}$} \\
    \bottomrule
    \end{tabular}}
    \end{center}
\end{wraptable}The bits/dim scores of RealNVP with conventional multi-scale architecture and RealNVP with LCMA are given in Table ~\ref{table:log_likelihood}. It can be observed that the density estimation results using LCMA is in all cases better in comparison to the baseline. We observed that the improvement for CelebA is relatively high as compared to natural image datasets. This observation was expected as facial features often contain high redundancy and the flow model learns to put more importance (reflected in terms of high log-det) on selected dimensions that define the facial features. Proposed LCMA exposes such dimensions to more flow layers, making them more expressive and hence the significant improvement in BPD is observed. The improvement in bits/dim is less for natural image datasets because of the high variance among features defining them, which has been the challenge with image compression algorithms. Note that the improvement in density estimation is always relative to the original flow architecture (RealNVP in our case) over which we use our proposed LCMA, as we do not alter any architecture other than the dimension factorization method. 

\begin{wraptable}[7]{r}{0.5\textwidth}
\vspace{-4mm}
\caption{Density estimation results (in bits/dim) for various flow models with LCMA on CIFAR-10}
\vspace{-4mm}
    \label{table:all_flows}
    \begin{center}
    \resizebox{0.49\textwidth}{!}{%
    \begin{tabular}{p{3.2cm} | p{1.3cm} | p{0.8cm} | p{1.4cm} |p{1.3cm}}
    \toprule
    \centering
    Type of Multi -scale Architecture (MA) & RealNVP & Glow & i-ResNet & Residual Flows
    \\\midrule
    \centering
    Conventional MA & \makecell{$3.49$} & \makecell{$3.35$} & \makecell{$3.45$} & \makecell{$3.28$} 
    \\\midrule
    \centering
    LCMA   & \makecell{$\mathbf{3.43}$} & \makecell{$\mathbf{3.31}$} & \makecell{$\mathbf{3.40}$}  & \makecell{$\mathbf{3.25}$}
    \\\bottomrule
    \end{tabular}}
    \end{center}
\end{wraptable}The quantitative results of LCMA implementation for several state-of-the-art flow models with CIFAR-10 dataset is summarized in Table ~\ref{table:all_flows}. The density estimation score for flow with LCMA outperforms the same flow with conventional multi-scale architecture. We also achieve state-of-the-art density estimation results for CIFAR-10 dataset with LCMA implementation for Residual Flows.
\begin{figure*}[h]
    { \centering                             
        \subfigure[\label{fig:multiscale_orig}Examples from the dataset]{\includegraphics[width=0.3\textwidth]{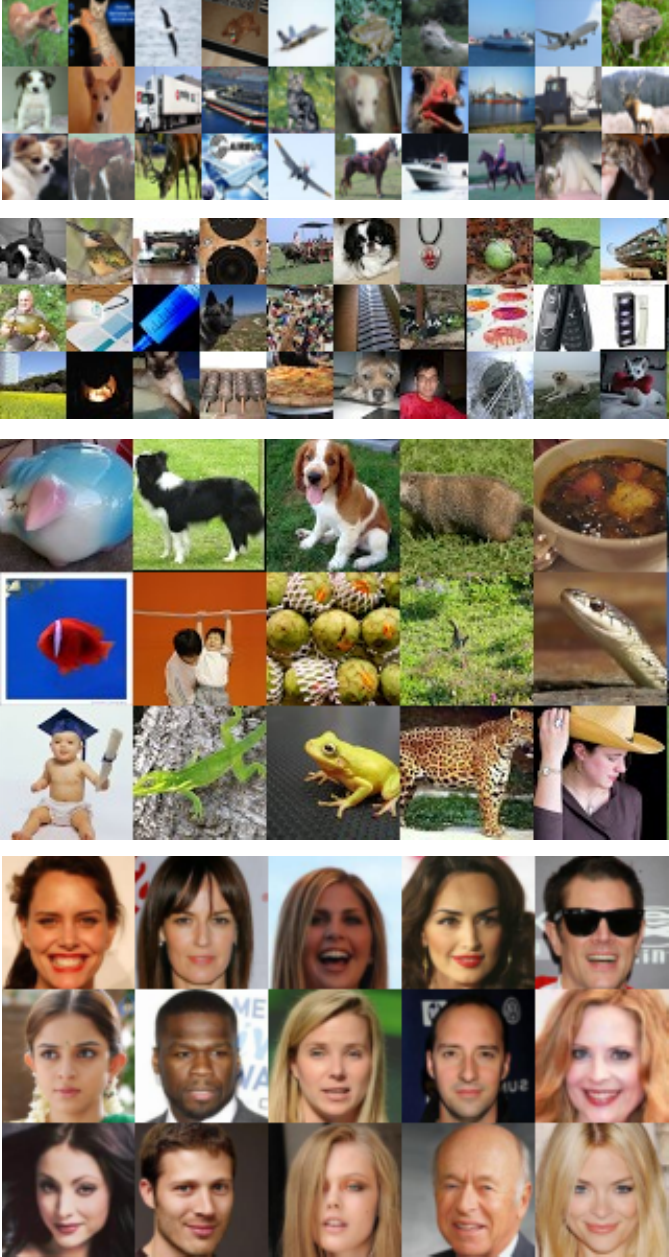}}
        \hspace{0.01\textwidth}
        \centering
        \subfigure[\label{fig:multiscale_realnvp}Samples from trained RealNVP \citep{dinh2016real}]{\includegraphics[width=0.3\textwidth]{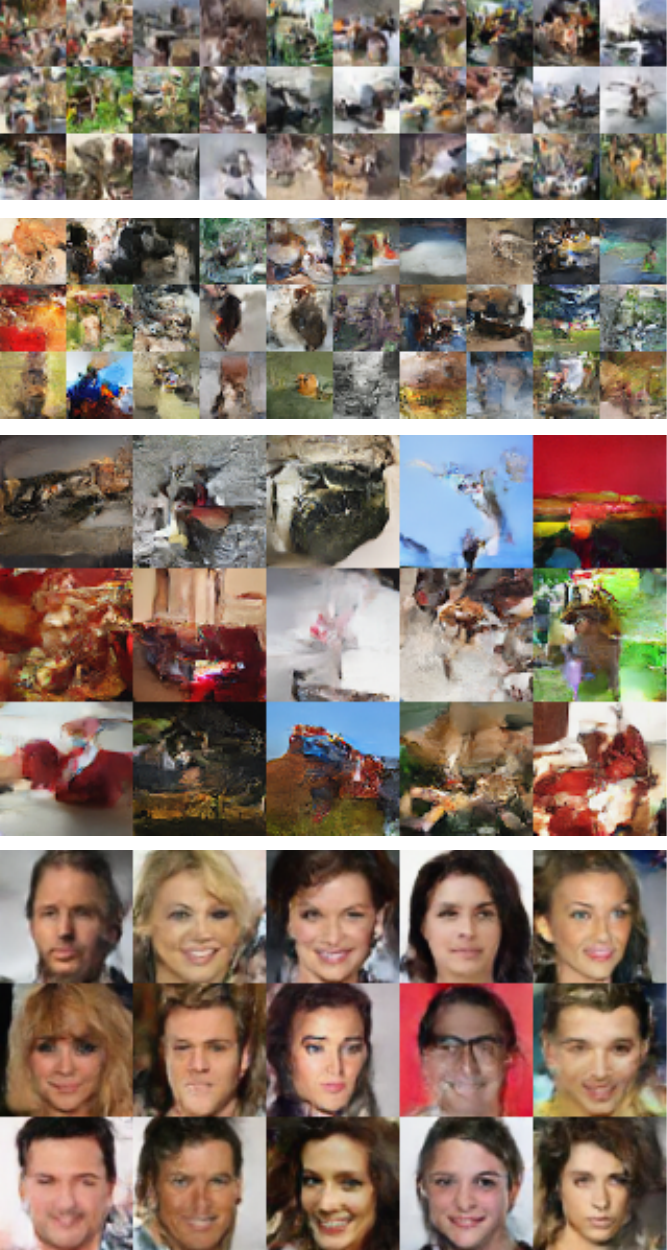}}
        \hspace{0.01\textwidth}
        \centering
        \subfigure[\label{fig:multiscale_sample}Samples from trained RealNVP flow model with LCMA]{\includegraphics[width=0.3\textwidth]{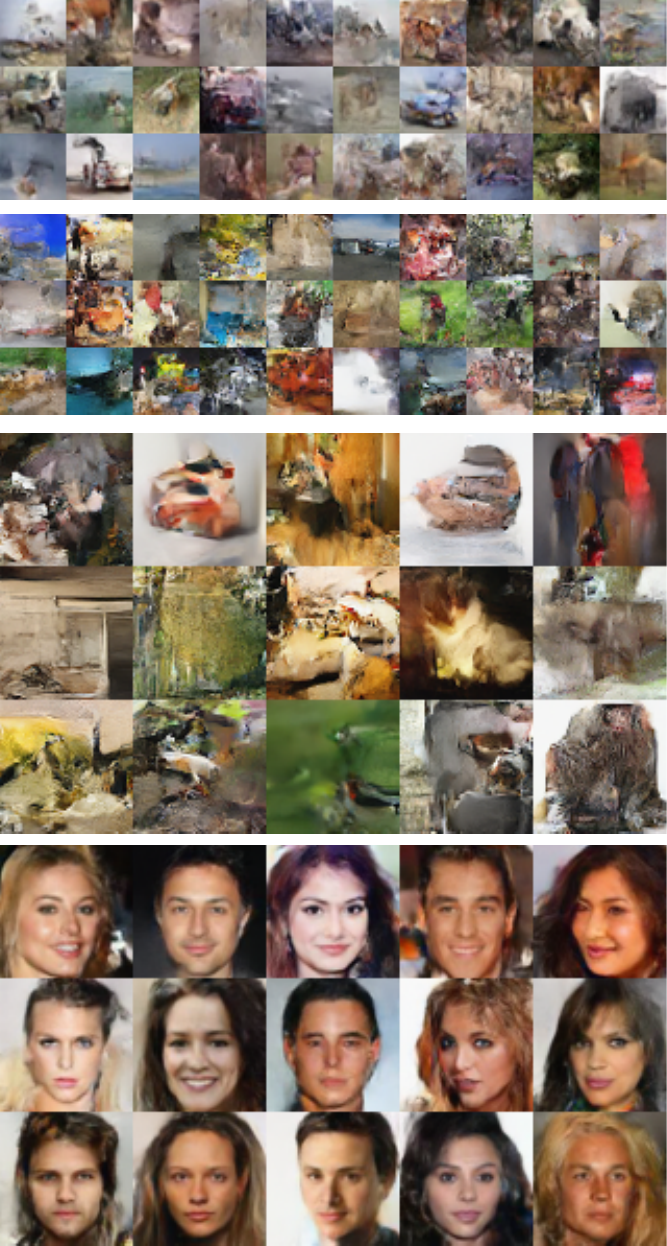}}
    }
    \caption{Samples from RealNVP \citep{dinh2016real} and RealNVP flow model with proposed LCMA trained on different datasets. The datasets shown in this figure are in order: CIFAR-10, Imagenet($32\times 32$), Imagenet ($64\times 64$) and CelebA (without low-temperature sampling).}
    \label{fig:data_n_samples}
    \vspace{-6mm}
\end{figure*}
\subsection{Qualitative Comparison}
\begin{wrapfigure}[11]{r}{0.4\textwidth}
\vspace{-4mm}
  \begin{center}
    \includegraphics[width=0.39\textwidth]{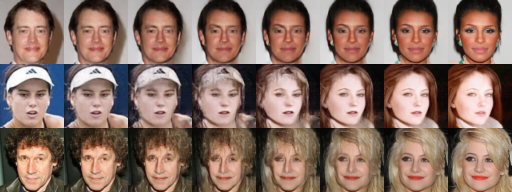}
  \end{center}
  \caption{Smooth linear interpolations in latent space between two images from CelebA. The intermediate samples perceptibly resemble synthetic faces.}
  \label{fig:interpolations}
\end{wrapfigure}For LCMA implementation, we introduced local max and min pooling operations (to preserve spatiality) on log-det heuristic to decide which dimensions to be gaussianized early (Section ~\ref{sensitive}). Fig. ~\ref{fig:multiscale_orig} shows samples from original datasets, Fig. ~\ref{fig:multiscale_realnvp} shows the samples from a trained RealNVP flow model with conventional multi-scale architecture and Fig. ~\ref{fig:multiscale_sample} shows the samples from RealNVP with LCMA, trained on various datasets. The finer facial details such as hair styles, eye-lining and facial folds in Celeba samples generated from RealNVP with LCMA were perceptually better than the baseline. The global feature representation observed is similar to that in RealNVP, as the flow architecture was kept the same. The background for natural images such as Imagenet were constructed at par with the original flow model. As it has been observed, for flow models, the latent space holds knowledge about the feature representation in the data. We performed linear interpolations in latent space to ensure its efficient construction. The interpolations observed (Fig.~\ref{fig:interpolations}) were smooth, with intermediate samples perceptibly resembling synthetic faces, signifying the efficient construction of latent space.

\subsection{Ablation Study}\label{sec:ablation}
\begin{wraptable}[10]{r}{0.65\textwidth}
\vspace{-4mm}
    \caption{Ablation study results for multi-scale architectures with various factorization methods trained on CelebA dataset}
    \vspace{-4mm}
    \label{table:ablation}
    \begin{center}
    \begin{small}
    \resizebox{0.64\textwidth}{!}{%
    \begin{tabular}{p{2.5cm} | p{2.1cm} | p{3.1cm} | p{2.1cm} | p{3.1cm}}
    \toprule
    \centering
    Evaluations   & \makecell{Fixed Random\\ Permutation} & \makecell{Early gaussianization of\\ \textit{high} log-det dimensions} & \makecell{RealNVP} & \makecell{Early gaussianization of\\ \textit{low} log-det dimensions}
    \\\midrule
    \centering
    Quantitative (BPD) & \makecell{$3.05$} & \makecell{$3.10$} & \makecell{$3.02$} & \makecell{$2.71$}
    \\\midrule
    \centering
    Qualitative    & \makecell{{\includegraphics[width=0.15\textwidth]{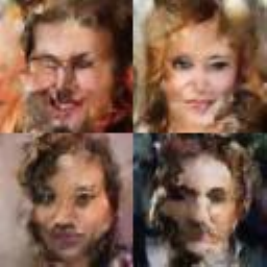}}} & \makecell{{\includegraphics[width=0.15\textwidth]{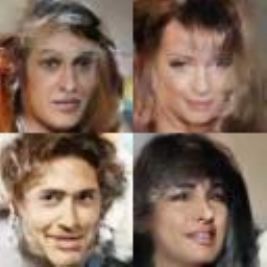}}} & \makecell{{\includegraphics[width=0.15\textwidth]{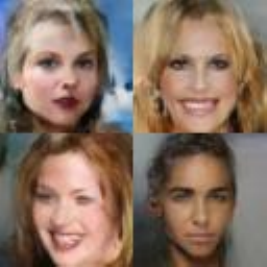}}} & \makecell{{\includegraphics[width=0.15\textwidth]{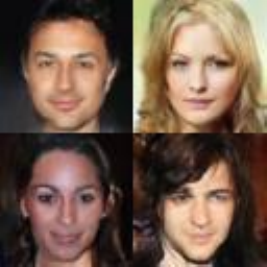}}} 
    \\\bottomrule
    \end{tabular}}
    \end{small}
\end{center}
\end{wraptable}
We perform two types of ablation studies to compare LCMA with other methods for dimension factorization in a multi-scale architecture. In our first study, we consider 4 variants, namely fixed random permutation (Case 1), multiscale architecture with early gaussianization of high log-det dimensions (Case 2), factorization method with checker-board and channel splitting as introduced in RealNVP (Case 3) and multiscale architecture with early gaussianization of low log-det dimensions, which is our proposed LCMA (Case 4). In fixed random permutation, we randomly partition the tensor into two halves, with no regard to the spatiality or log-det score. In case 2, we do the reverse of LCMA, and early gaussianize the high log-det variables at each layer. The bits/dim score and generated samples for each of the methods are given in Table ~\ref{table:ablation}. As expected from an information theoretic perspective, gaussianizing high log-det variables early provides the worst density estimation, as the model could not capture the high amount of important information. Comparing the same with fixed random permutation, the latter has better score as the probability of a high log-det variable being gaussianized early reduces to half, and it gets further reduced with RealNVP due to channel-wise and checkerboard splitting. LCMA has the best score among all methods, as the variables more sensitive to changes in input (hence carrying more information) are exposed to more flow layers. Fixed random permutation has the worst quality of sampled images, as the spatiality is lost during factorization. The sample quality improves for Case 2 and RealNVP. The sampled images are perceptually best for LCMA.

\begin{wraptable}[12]{r}{0.3\textwidth}
\vspace{-4mm}
\caption{Ablation study results for permuting factorization of high/low log-det dims}
\vspace{-4mm}
\label{tab:ablation_perm}
\vskip 0.15in
\begin{center}
\begin{small}
\resizebox{0.29\textwidth}{!}{%
\begin{tabular}{p{3.2cm} | p{1cm}}
\toprule
Permutation of high/low log-det dimensions & Bits/dim \\
\midrule
High-High-High    & 3.10 \\
High-High-Low & 3.09\\
High-Low-High    & 3.07\\
High-Low-Low    & 3.05\\
Low-High-High     & 3.00\\
Low-High-Low      & 2.92\\
Low-Low-High      & 2.79\\
Low-Low-Low (LCMA)   & 2.71\\
\bottomrule
\end{tabular}}
\end{small}
\end{center}
\vskip -0.1in
\end{wraptable}We perform a second ablation study to reconfirm that early gaussianization of high log-det dimensions has a deteriorating effect on the density estimation score. The flow model in our experiment has 3 layers where dimensions splitting is being performed. We consider all permutations of early gaussianization of high/low log-det dimensions at each of the 3 layers. The density estimation scores for all $2^3$ permutations trained on CelebA dataset are presented in Table ~\ref{tab:ablation_perm}. The best score corresponds to early gaussianization of low log-det dimensions at each flow layer (proposed LCMA), and the score deteriorates with permutations involving early gaussianization of high log-det dimensions at any flow layer. Summarizing, LCMA outperforms multi-scale architectures based on other factorization methods, as it improves density estimation scores and generates qualitative samples.

\section{Conclusions}
We proposed a novel multi-scale architecture for generative flows which employs a data-dependent splitting based on the individual contribution of dimensions to the log-likelihood. Implementations of the proposed method for several flow models such as RealNVP \citep{dinh2016real}, Glow\citep{kingma2018glow}, i-ResNet\citep{behrmann2018invertible} and Residual Flows \citep{chen2019residual} were presented. Empirical studies conducted on benchmark image datasets validate the strength of our proposed method, which improves log-likelihood scores and is able to generate qualitative samples. Ablation study results confirm the power of LCMA over other options for dimension factorization. A line of future work can be to design/learn a masking scheme for factorization online during flow training (or possibly a parallel training process), while preserving flow properties.
\newpage
\section*{Broader Impact}
Generative models in general require higher amounts of computations in order to be trained with a dense network and/or larger amounts of data. This becomes a big challenge for real-world high-dimensional data, especially when the model itself has constraints of bijectivity, e.g. generative flows. Proposed Likelihood Contribution based Multi-scale Architecture (LCMA) provides an avenue to evaluate the importance of individual dimensions present in the data and use it to make flow based generative models computationally efficient and at the same time powerful density estimators.

We envision proposed research will open up discussions on making flow based generative models computationally efficient, so as to allow a broad section of disciplines reap benefits of it for their big-data applications while remaining within their computational budgets. Particularly this will make researchers with limited budget get access and contribute to the generative AI world.  Optimized flow models can also be used alongside real-world applications such as energy/smart buildings ~\citep{zou2019consensus,zou2019wifi,zou2019machine,konstantakopoulos2019design,chen2021enforcing,periyakoil2021environmental,periyakoil2020understanding,das2019novel,das2020occupants,liu2018personal,liu2019personal,donti2021machine,jin2018biscuit,das2021unsupervised}, healthcare~\citep{das2021conditional}, vision~\citep{das2021cdcgen,yue2021multi,gong2019dlow} etc. Additionally, since we focus on quantifying the sensitivity of each data dimension to the input data variations, we expect our method will potentially be used in data compression, semi-supervised learning and to deal with data bias, which are important components in dealing with AI for societal systems.  

There still are directions where the proposed research has room to be more efficient (which we also mention as future work). Currently, proposed dimension splitting occurs offline (not during the actual training process), since doing it otherwise would violate \textit{change of variables} formula, a general principle for flow training. A design of dimension splitting which directly/indirectly occurs online (during the training process) and is in line with flow training principles can make the method even more computationally efficient, hence more accessible to researchers with limited budget.
\section*{Acknowledgement}
This research was funded by the Republic of Singapore's National Research Foundation through a grant to the Berkeley Education Alliance for Research in Singapore (BEARS) for the Singapore-Berkeley Building Efficiency and Sustainability in the Tropics (SinBerBEST) Program. BEARS has been established by the University of California, Berkeley as a center for intellectual excellence in research and education in Singapore.
\bibliographystyle{abbrvnat}
\bibliography{references.bib}
\end{document}